%% file: main.tex
\documentclass[11pt,a4paper]{article}
\usepackage[hyperref]{acl2020}
\usepackage[utf8]{inputenc}
\usepackage{times}  
\usepackage{helvet}  
\usepackage{courier}  
\usepackage{url}  
\usepackage{graphicx}  
\usepackage{latexsym}
\usepackage{multirow}
\usepackage{tikz,tikz-qtree}
\usepackage{amsmath,amsfonts,amssymb}
\usepackage[linesnumbered,ruled]{algorithm2e}
\usepackage{algorithmic}
\usepackage{verbatim}
\usepackage{CJK}
\usepackage{xcolor}
\usepackage{fancybox}
\usepackage{todonotes}
\usepackage{pifont}
\usepackage{booktabs}
\usepackage{enumitem}
\usepackage[caption=false]{subfig}

\newcommand{\eat}[1]{\ignorespaces}
\newcommand{\cut}[1]{}

\newcommand\sopk{{\cal{S}}^{0}_{k}}
\newcommand\spk{{\cal{S}}_{k}}
\newcommand\spn{{\cal{S}}_{n}}

\newcommand{\emptyspan}{\textsc{null}}

\usepackage{todonotes}
\newcounter{note}
\newcounter{chenghao}
\newcommand{\chenghao}[1]{%
\refstepcounter{chenghao}%
{%
\todo[color=green, size=\footnotesize]{%
[\textbf{chenghao:\thechenghao}] #1}%
}}%

\newcounter{mw}
\newcommand{\mw}[1]{%
\refstepcounter{mw}%
{%
\todo[color=orange, size=\footnotesize]{%
[\textbf{mw:\mw}] #1}%
}}%

\newcounter{kristina}
\input{Definitions.tex}

\usepackage{url}

\title{ Probabilistic Assumptions Matter: Improved Models for Distantly-Supervised Document-Level Question Answering}
\author{Hao Cheng$^{*}$, Ming-Wei Chang$^\dagger$,  Kenton Lee$^\dagger$,  Kristina Toutanova$^\dagger$ \\
\\ $^*$Microsoft Research\\
\texttt{\large chehao@microsoft.com}\\\\
$^\dagger$Google Research \\
\texttt{\large \{mingweichang, kentonl, kristout\}@google.com} \\
}
 \aclfinalcopy
\begin{document}
\maketitle

\begin{abstract}
We address the problem of extractive question answering using document-level distant supervision, pairing questions and relevant documents with answer strings.
We compare previously used probability space and distant supervision assumptions (assumptions on the correspondence between the weak answer string labels and possible answer mention spans). We show that these assumptions interact, and that different configurations provide complementary benefits.
We demonstrate that a multi-objective model can efficiently combine the advantages of multiple assumptions and outperform the best individual formulation. 
Our approach outperforms previous state-of-the-art models by 4.3 points in F1 on TriviaQA-Wiki and 1.7 points in Rouge-L on  NarrativeQA summaries.\footnote{Based on the TriviaQA-Wiki leaderboard, our approach was the SOTA when this work was submitted on Dec 04, 2019.}

\end{abstract}


\input{sections/01_introv2.tex}
\input{sections/02_model_description.tex}
\input{sections/03_v2_distant_supervision.tex}

\input{sections/04_optimization_and_inference.tex}

\input{sections/05_experiment.tex}

\input{sections/06_analysis.tex}

\input{sections/07_related.tex}

\input{sections/08_conclusion.tex}

\section*{Acknowledgement}
Some of the ideas in this work originated from Hao Cheng's internship with Google Research. We would like to thank Ankur Parikh, Michael Collins, and William Cohen for discussion and detailed feedback on this work, as well as other members from the Google Research Language team and the anonymous reviewers for valuable suggestions. We would also like to thank Sewon Min for generously sharing the processed data and evaluation script for NarrativeQA.

\bibliography{textandknowledge}
\bibliographystyle{acl_natbib}

\end{document}

%% file: sections/01_introv2.tex
\section{Introduction}
\label{sec:intro}


\begin{figure}[th!]
    \centering
    \includegraphics[width=0.48\textwidth]{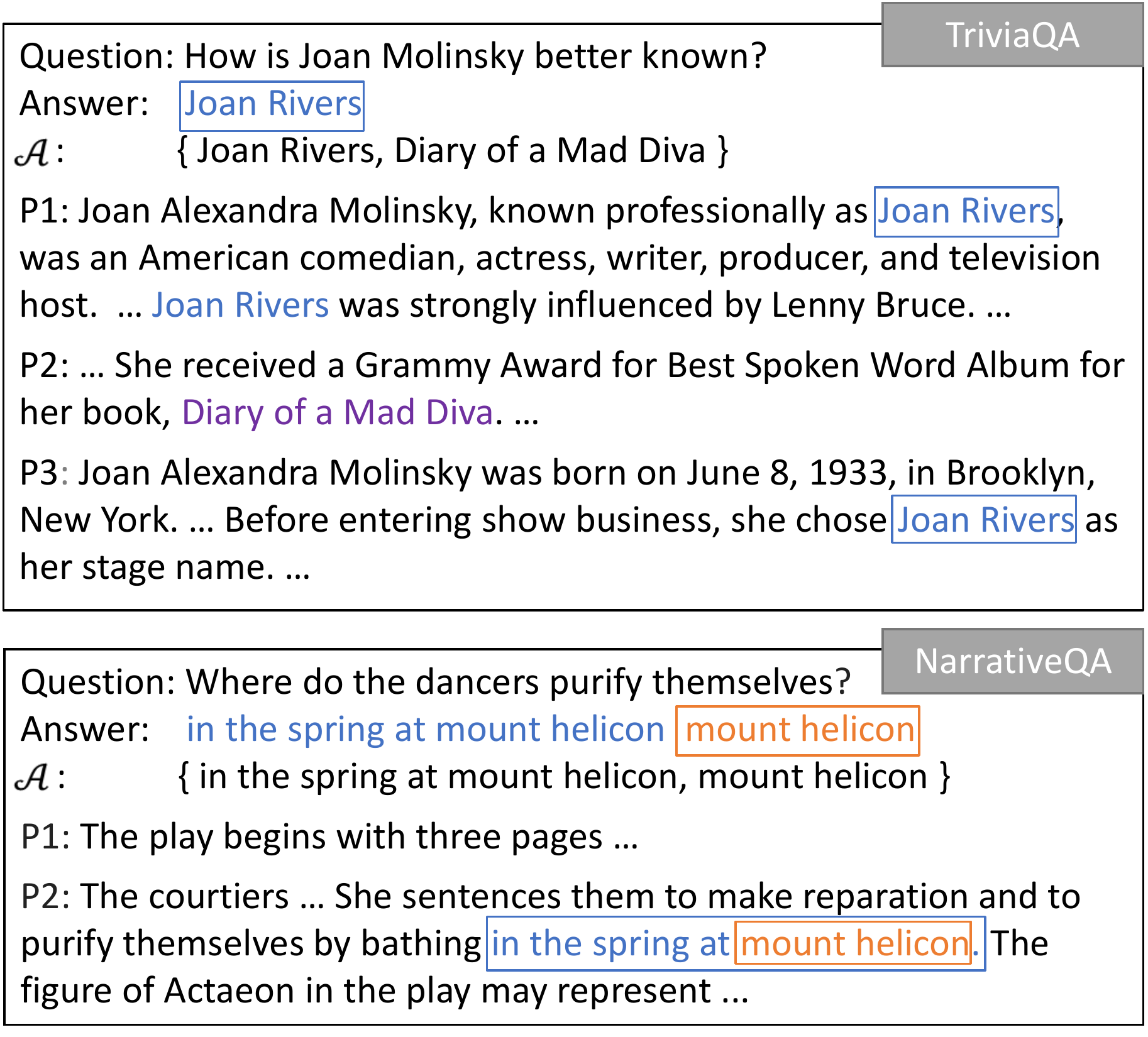}
    \caption{\small{TriviaQA and NarrativeQA examples.
    In the TriviaQA example, there are three occurrences of the original answer string ``Joan Rivers'' ({blue}),
    and one alternate but incorrect alias  ``Diary of a Mad Diva'' (purple). 
    Only two ``Joan Rivers'' mentions (shown in blue boxes) support answering the question.
    In the NarrativeQA example, there are two answer stings in $\cal{A}$: 
    ``in the spring at mount helicon'' (blue) and ``mount helicon'' (orange),
    with the latter being a substring of the former.
    Both mentions in P2 are correct answer spans.
    }
    }
    \label{fig:noisy_sample}
\end{figure}

Distant supervision assumptions have enabled the creation of large-scale datasets that can be used to train fine-grained extractive short answer question answering (QA) systems. 
One example is TriviaQA~\cite{joshi-EtAl:2017:Long}. There the authors utilized a pre-existing set of Trivia question-answer string pairs and coupled them with relevant documents, such that, with high likelihood, the documents support answering the questions (see \autoref{fig:noisy_sample} for an illustration). Another example is the NarrativeQA dataset~\cite{Kocisky2018TACL-narrativeqa}, where crowd-sourced abstractive answer strings were used to weakly supervise answer mentions in the text of movie scripts or their summaries. In this work, we focus on the setting of document-level extractive QA, where distant supervision is specified as a set $\mathcal{A}$ of answer strings for an input question-document pair. 

Depending on the data generation process, the properties of the resulting supervision from the sets $\mathcal{A}$ may differ.  For example, the provided answer sets in TriviaQA include aliases of original trivia question answers, aimed at capturing semantically equivalent answers but liable to introducing semantic drift. 
In \autoref{fig:noisy_sample},
the possible answer string ``Diary of a Mad Diva'' is related to ``Joan Rivers'', but is not a valid answer for the given question.

On the other hand, the sets of answer strings in NarrativeQA are mostly valid since they have high overlap with human-generated answers for the given question/document pair. As shown in \autoref{fig:noisy_sample},  ``in the spring at mount helicon'' and ``mount helicon'' are both valid answers with relevant mentions. In this case, the annotators chose answers that appear verbatim in the text but in the more general case, noise may come from partial phrases and irrelevant mentions.

While distant supervision reduces the annotation cost, increased coverage often comes 
with increased noise (e.g., expanding entity answer strings with aliases improves coverage but also increases noise). Even for  fixed  document-level distant supervision in the form of a set of answers $\mathcal{A}$, different interpretations of the partial supervision lead to different points in the coverage/noise space and their relative performance is not well understood.



This work systematically studies
methods for learning and inference with document-level distantly supervised extractive QA models. 
Using a 
BERT~\cite{devlin-etal-2019-bert} joint question-passage encoder, we study the compound impact of:

\begin{itemize}[leftmargin=*,topsep=0pt,itemsep=-1ex,partopsep=1ex,parsep=1ex]
    \item \textbf{Probability space} (\S\ref{sec:approach}):
    ways to define the model's probability space 
    based on independent paragraphs or whole documents.
    
    \item \textbf{Distant supervision assumptions} (\S\ref{sec:distant_supervision}): ways to translate the supervision from possible strings $\mathcal{A}$ to possible locations of answer mentions in the document. 
    
    \item \textbf{Optimization and inference} (\S\ref{sec:opt_and_infer}): ways to define corresponding 
    training objectives
    (e.g.\ Hard EM as in \newcite{Min-2019-EMNLP-hardem} vs.\ Maximum Marginal Likelihood) and make answer string predictions during inference (Viterbi or marginal inference).
\end{itemize}

We show that the choice of probability space puts constraints on the distant supervision assumptions that can be captured,
and that all three choices interact, leading to large differences in performance.
Specifically, we provide a framework for understanding different distant supervision assumptions and the corresponding trade-off among the coverage, quality and strength of distant supervision signal.
The best configuration depends on the properties of the possible annotations $\mathcal{A}$ and is thus data-dependent.
Compared with recent work also using BERT representations, our study show that the model with most suitable probabilistic treatment achieves large improvements of 4.6 F1 on TriviaQA and 1.7 Rouge-L on NarrativeQA respectively.
Additionally, we design an efficient multi-loss objective that can combine the benefits of different formulations, leading to significant improvements in accuracy, surpassing the best previously reported results on the two studied tasks. Results are further strengthened by transfer learning from fully labeled short-answer extraction data in SQuAD 2.0~\cite{rajpurkar-etal-2018-know}, leading to a final state-of-the-art performance of 76.3 F1 on TriviaQA-Wiki and 62.9 on the NarrativeQA summaries task.\footnote{The code is available at \url{https://github.com/hao-cheng/ds_doc_qa}}

%% file: sections/02_model_description.tex
\section{Probability Space}
\label{sec:approach}
\begin{figure}
    \centering
    \includegraphics[width=0.48\textwidth]{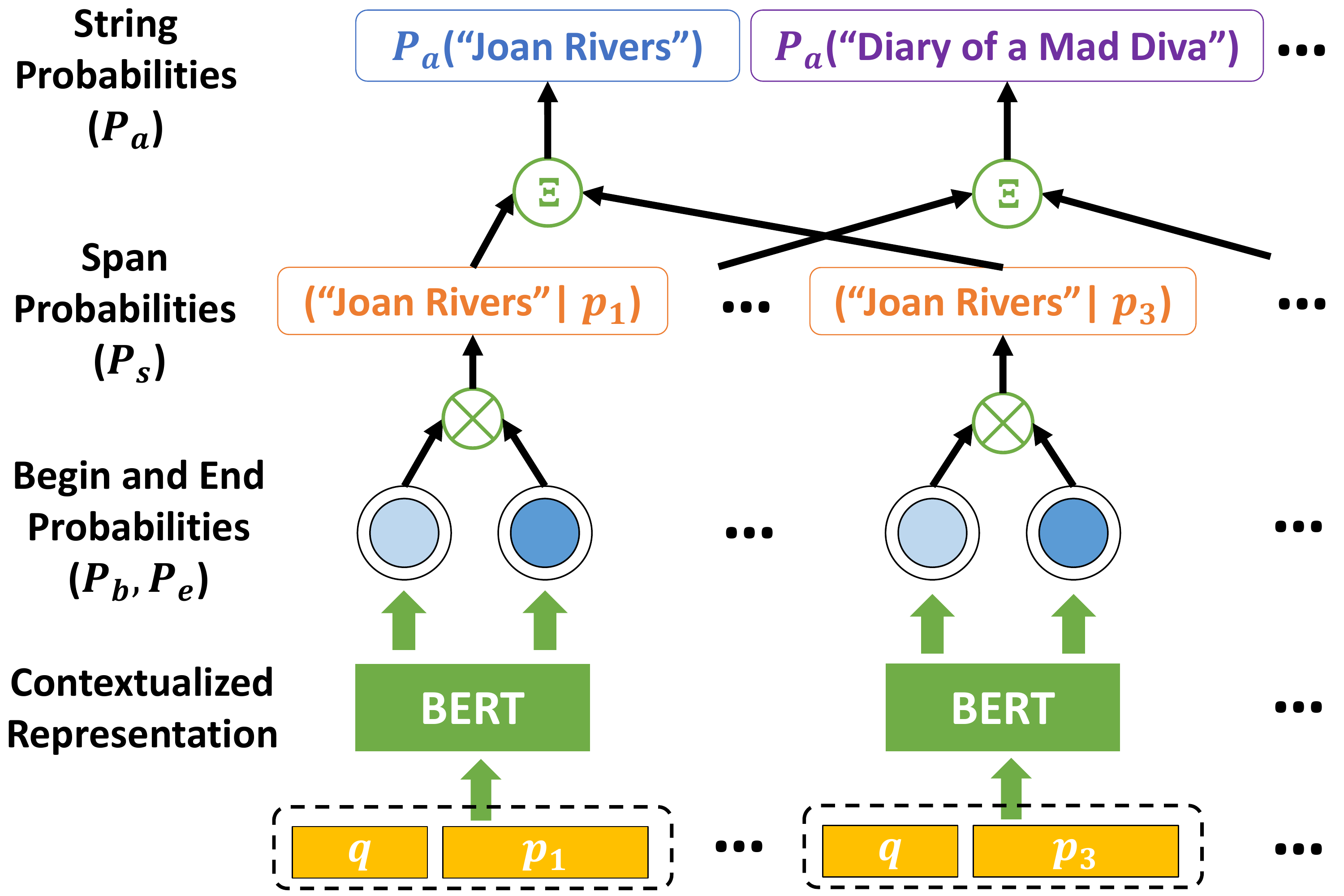}
    \caption{The document-level QA model as used for test-time inference.
    The lower part is a BERT-based paragraph-level answer scoring component, and the upper part illustrates the probability aggregation across answer spans sharing the same answer string.
    $\Xi$ refers to either a sum or a max operator.
    In the given example, ``John Rivers'' is derived from two paragraphs.
    }
    \label{fig:model_overview}
\end{figure}

Here, we first formalize both paragraph-level and document-level models, which have been previously used for document-level extractive QA. Typically, paragraph-level models consider each paragraph in the document independently, whereas document models integrate some dependencies among paragraphs.

To define the model, we need to specify the probability space, consisting of a set of possible \emph{outcomes} and a way to assign \emph{probabilities} to individual outcomes.
For extractive QA, the probability space outcomes consist of token positions of answer mention spans.

The overall model architecture is shown in \autoref{fig:model_overview}.
We use BERT~\cite{devlin-etal-2019-bert} to derive representations of document tokens.
As is standard in state-of-the-art extractive QA models \cite{devlin-etal-2019-bert,lee-etal-2019-latent,Min-2019-EMNLP-hardem}, 
the BERT model is used to encode a pair of a given question with one paragraph from a given document into neural text representations.
These representations are then used to define scores/probabilities of possible answer begin and end positions, which are in turn used to define probabilities over possible answer \emph{spans}. Then the answer string probabilities can be defined as the aggregation over all possible answer spans/mentions.

In the following, we show that
paragraph-level and document-level models differ only in the space of possible outcomes and the way of computing answer span probabilities from answer position begin and end scores.

\paragraph
{Scoring answer begin and end positions}
Given a question $q$ and a document $d$ consisting of $K$ paragraphs $p_1, \ldots, p_K$,
the BERT encoder produces contextualized representations for each question-paragraph pair $(q, p_k)$.
Specifically, for each token position $i^k$ in 
$p_k$,
the final hidden vector $\hvec^{(i, k)} \in \RR^d$ 
is used as the contextualized token embedding, where $d$ is the vector dimension.

The span-begin score is computed as 
$s_b(i^k) = \wvec_b^T \hvec^{(i, k)}$ using a weight vector
$\wvec_b \in \RR^d$.
The span-end score $s_e(j^k)$ is defined in the same way.
The probabilities for a start position $i^k$ and an end position $j^k$ are
\begin{eqnarray}
\label{eqn:span_begin_prob}
P_b(i^k) = {\exp(s_b(i^k)) \over Z_b},\\
\label{eqn:span_end_prob}
P_e(j^k) = {\exp(s_e(j^k)) \over Z_e},
\end{eqnarray}
where $Z_b, Z_e$ are normalizing factors, depending on the probability space definition (detailed below).
The probability of an answer span from $i^k$ to $j^k$ is defined as 
$ P_s(i^k, j^k) = P_b(i^k) P_e(j^k) $.

The partition functions $Z_b$ and $Z_e$ depend on whether we use a paragraph-level or document-level probability space.


%


\paragraph
{Paragraph-level model} In paragraph level models, we assume that for a given question against a document $d$, each of its paragraphs $p_1,\ldots,p_K$ independently selects a pair of answer positions $(i^k,j^k)$, which are the begin and end of the answer from paragraph $p_k$. In the case that $p_k$ does not support answering the question $q$, special ${\emptyspan}$ positions are selected (following the SQuAD 2.0 BERT implementation\footnote{\smaller \url{https://github.com/google-research/bert}}). 
Thus, the set of possible outcomes $\Omega$ in the paragraph-level probability space is the set of lists of begin/end position pairs, one from each paragraph: $\{[(i^1,j^1),\ldots, (i^K,j^K)]\}$, where $i^k$ and $j^k$ range over positions in the respective paragraphs.

The answer positions in different paragraphs are independent, and the probability of each paragraph's answer begin and end is computed by normalizing over all possible positions in that paragraph, i.e.,
\begin{align}
\label{eqn:par_begin_z}
    Z_b^k = \sum_{i \in \mathcal{I}^k \cup \{ \emptyspan \} } \exp (s_b(i)), \\
\label{eqn:par_end_z}
Z_e^k = \sum_{j \in \mathcal{I}^k \cup \{ \emptyspan \}} \exp (s_e(j)),
\end{align}
where $\mathcal{I}^k$ is the set of all positions in the paragraph $p_k$. 
The probability of an answer begin at $i^k$ is 
$ P_b(i^k) = \exp (s_b(i^k)) / {Z_b}^k $ 
and the probability of an end at $j^k$ is defined analogously.
The probability of a possible answer position assignment for the document $d$ is then defined as
$P([(i^1,j^1),\ldots, (i^K,j^K)]) = \prod_{k}{P_b(i^k)P_e(j^k)}.$

As we can see from the above definition, due to the independence assumption, models using paragraph-level normalization do not learn to directly calibrate candidate answers from different paragraphs against each other.

\paragraph
{\bf Document-level model}
In document-level models, we assume that for a given question against document $d$, a single answer span is selected (as opposed to one for each paragraph in the paragraph-level models).\footnote{In this paper, we focus on datasets where the document is known to contain a valid answer. It is straightforward to remove this assumption and consider document-level ${\emptyspan}$ for future work.}
Here, the possible positions in all paragraphs are a part of a joint probability space and directly compete against each other.

In this case, $\Omega$ is the set of token spans $\{(i,j)\}$, where $i$ and $j$ are the begin and end positions of the selected answer. 
The normalizing factors 
are therefore aggregated over all paragraphs, i.e.,
\begin{align}
\label{eqn:doc_begin_z}
Z_b^* = \sum_{k=1}^K \sum_{i \in \mathcal{I}^k} \exp (s_b(i)),  \\
\label{eqn:doc_end_z}
Z_e^* = \sum_{k=1}^K \sum_{j \in \mathcal{I}^k} \exp (s_e(j)).
\end{align}
Compared with \eqref{eqn:par_begin_z} and \eqref{eqn:par_end_z}, since there is always a valid answer in the document for the tasks studied here, \emptyspan\ is not necessary for document-level models and thus can be excluded from the inner summation of \eqref{eqn:doc_begin_z} and \eqref{eqn:doc_end_z}.
The probability of a possible outcome, \ie an answer span, is
$P(i,j) = \exp (s_b(i) + s_e(j)) / (Z_b^* Z_e^*).$

%% file: sections/03_v2_distant_supervision.tex
\section{Distant Supervision Assumptions}
\label{sec:distant_supervision}

%
There are multiple ways to interpret the distant supervision 
signal from $\cal{A}$ as possible outcomes in our paragraph-level and document-level probability spaces, leading to corresponding training loss functions.
Although several different paragraph-level and document-level losses \cite{drqa,kadlec-etal-2016-text,clarkgardner,lin-etal-2018-denoising,Min-2019-EMNLP-hardem} have been studied in the literature, we want to point out that when interpreting the distant supervision signal, there is a tradeoff among multiple desiderata: 

\begin{itemize}[leftmargin=*,nolistsep]
    \item \textbf{Coverage}: 
    maximize the number of instances of relevant answer spans, which we can use to provide positive examples to our model.
    \item \textbf{Quality}: 
    maximize the quality of annotations by minimizing noise from irrelevant answer strings or   mentions.
    \item \textbf{Strength}: 
    maximize the strength of the signal by reducing  uncertainty and pointing the model more directly at correct answer mentions.
\end{itemize}
We introduce three assumptions (\texttt{H1, H2, H3}) for how the distant supervision signal should be interpreted, which lead to different tradeoffs among the desiderata above (see \autoref{tab:ds_hypo_and_tradeoffs}).
\begin{table}[]
    \centering
    \begin{tabular}{c|c|c|c}
    \toprule
                     & Coverage & Quality & Strength \\
        \midrule
         \texttt{H1} &  $\nearrow$ & $\searrow$ & $\nearrow$ \\
         \texttt{H2} &  $\longrightarrow$ & $\longrightarrow$ & $\longrightarrow$ \\
         \texttt{H3} &  $\searrow$ & $\nearrow$ & $\searrow$ \\
         \bottomrule
    \end{tabular}
    \caption{Distant supervision assumptions and their corresponding tradeoffs.
    ($\nearrow$) indicates highest value, ($\rightarrow$) medium, and ($\searrow$)  lowest value.}
    \label{tab:ds_hypo_and_tradeoffs}
\end{table}

We  begin with setting up additional useful notation.
Given a document-question pair $(d, q)$ and a set of answer strings $\cal{A}$, we define the set of $\cal{A}$-consistent token spans $\cal{Y}_{\cal{A}}$ in $d$ as follows:
for each paragraph $p_k$, span $(i^k,j^k) \in {\cal{Y}}^k_{\cal{A}}$ if and only if the string spanning these positions in the paragraph is in  $\cal{A}$. 
For paragraph-level models, if for paragraph $p_k$ the set ${\cal{Y}}^k_{\cal{A}}$ is empty, we redefine ${\cal{Y}}^k_{\cal{A}}$ to be $\{$\emptyspan$\}$.
Similarly, we define the set of $\cal{A}$-consistent begin positions ${{\cal{Y}}^k_{b,\cal{A}}}$ as the start positions of consistent spans: ${{\cal{Y}}^k_{b,\cal{A}}} = {\cup}_{(i,j) \in {\cal{Y}}^k_{\cal{A}}} \{i\} $. 
${{\cal{Y}}^k_{e,\cal{A}}}$ for $\cal{A}$-consistent end positions is defined analogously. In addition, we term an answer span $(i,j)$ \emph{correct} for question $q$, 
if its corresponding answer string is a correct answer to $q$,
and the context of the specific mention of that answer string from positions $i$ to $j$ entails this answer.
Similarly, we term an answer begin/end position \emph{correct} if there exists a correct answer span starting/ending at that position.

\begin{table*}[t]
    \centering
    \begin{tabular}{c|c|c}
    \toprule
    & Span-Based & Position-Based  \\
    \midrule
    \texttt{H1}
    & $\sum_{k \in \mathcal{K}} \sum_{(i^k, j^k) \in \mathcal{Y}^k_\mathcal{A}} \log P_s (i^k, j^k)$ 
    &  $\sum_{k \in \mathcal{K}} \sum_{i^k \in \mathcal{Y}^k_{b, \mathcal{A}}} \log P_b (i^k) +
        \sum_{k \in \mathcal{K}} \sum_{j^k \in \mathcal{Y}^k_{e, \mathcal{A}}} \log P_e (j^k) $ \\
    \texttt{H2}
    & $\sum_{k \in \mathcal{K}} \log \Xi_{(i^k, j^k) \in \mathcal{Y}^k_\mathcal{A}} P_s (i^k, j^k)$
    & $\sum_{k \in \mathcal{K}} \log \Xi_{i^k \in \mathcal{Y}^k_{b, \mathcal{A}}} P_b (i^k) + 
       \sum_{k \in \mathcal{K}} \log \Xi_{j^k \in \mathcal{Y}^k_{e, \mathcal{A}}} P_e (j^k)$ \\
    \texttt{H3}
    & $\log \Xi_{k \in \mathcal{K}} \Xi_{(i^k, j^k) \in \mathcal{Y}^k_\mathcal{A}} P_s (i^k, j^k)$
    & $\log \Xi_{k \in \mathcal{K}} \Xi_{i^k \in \mathcal{Y}^k_{b, \mathcal{A}}} P_b (i^k) + 
       \log \Xi_{k \in \mathcal{K}} \Xi_{j^k \in \mathcal{Y}^k_{e, \mathcal{A}}} P_e (j^k)$ \\
    \bottomrule
    \end{tabular}
    \caption{Objective functions for a document-question pair ($d, q$) under different distant supervision assumptions.
    $\Xi$ refers to $\sum$ and $\max$ for MML and HardEM, respectively.}
    \label{tab:ds_assumptions}
\end{table*}


\noindent
\textbf{H1: All $\cal{A}$-consistent answer spans are correct.}
While this assumption is evidently often incorrect (low on the quality dimension $\searrow$), especially for TriviaQA, as seen from \autoref{fig:noisy_sample}, it provides a large number of positive examples and a strong supervision signal (high on coverage $\nearrow$ and strength $\nearrow$). We include this in our study for completeness.

\texttt{H1} translates differently into possible outcomes for corresponding models depending on the probability space (paragraph or document).
Paragraph-level models select multiple answer spans, one for each paragraph, to form a possible outcome. 
Thus, multiple  $\cal{A}$-consistent answer spans can occur in a single outcome, as long as they are in different paragraphs.
For multiple $\cal{A}$-consistent answer spans in the same paragraph, these can be seen as mentions that can be selected with equal probability (e.g., by different annotators).
Document-level models select a single answer span in the document and therefore multiple $\cal{A}$-consistent answer spans can be seen as occurring in separate annotation events. \autoref{tab:ds_assumptions} shows in row one the log-probability of outcomes consistent with \texttt{H1}.
 
\noindent
\textbf{H2: Every positive paragraph has a correct answer in its $\cal{A}$-consistent set.}
Under this assumption, each paragraph with a non-empty set of $\cal{A}$-consistent spans (termed a \emph{positive} paragraph) has a correct answer. As we can see from the TriviaQA example in \autoref{fig:noisy_sample}, this assumption is correct for the first and third paragraph, but not the second one, as it only contains a mention of a noisy answer alias. This assumption  has medium coverage ($\rightarrow$), as it generates positive examples from multiple paragraphs but does not allow multiple positive mentions in the same paragraph. It also decreases noise (higher quality $\rightarrow$) (e.g. does not claim that all the mentions of ``Joan Rivers'' in the first paragraph support answering the question). The strength of the supervision signal is weakened ($\rightarrow$) relative to \texttt{H1}, as now the model needs to figure out which of the multiple ${\cal{A}}$-consistent mentions in each paragraph is correct.

\texttt{H2} has two variations: correct \emph{span}, assuming that one of the answer spans $(i^k,j^k)$ in  ${\cal{Y}}^k_{\cal{A}}$ is correct, and correct \emph{position}, assuming that the paragraph has a correct answer begin position from ${{\cal{Y}}^k_{b,\cal{A}}}$ and a correct answer end position from ${{\cal{Y}}^k_{e,\cal{A}}}$, but its selected answer span may not necessarily belong to ${\cal{Y}}^k_{\cal{A}}$.
For example, if ${\cal{A}}$ contains 
$\{ abcd$, $bc \}$,
then 
$abc$ would have 
correct begin and end,
but not be a correct span. 
It does not make sense for modeling to assume the paragraph has correct begin and end positions instead of a correct answer span 
(i.e., we don't really want to get inconsistent answers like $abc$ above),
but given that our probabilistic model assumes independence of begin and end answer positions, it may not be able to learn well with span-level weak supervision.
Some prior work~\cite{clarkgardner} uses an \texttt{H2} position-based distant supervision assumption with a pair-paragraph model akin to our document-level ones.
\citet{lin-etal-2018-denoising} use an
\texttt{H2} span-based distant supervision assumption.
The impact of position vs.\ span-based modeling of the distant supervision is not well understood.
As we will see in the experiments, for the majority of settings, position-based weak supervision is more effective than span-based for our model.


For paragraph-level and document-level models,  \texttt{H2} corresponds differently to possible outcomes. For paragraph models, one outcome can select answer spans in all positive paragraphs and \emptyspan\ in negative ones. For document-level models, we view answers in different paragraphs as outcomes of multiple draws from the distribution. The identity of the particular correct span or begin/end position is unknown, but we can compute the probability of the event comprising the consistent outcomes. Table~\ref{tab:ds_assumptions} shows the log-probability of the outcomes consistent with \texttt{H2} in row two (right for span-based and left for position-based interpretation, when plugging in $\sum$ for  $\Xi$).


\noindent
{\bf H3: The document has a correct answer in its $\cal{A}$-consistent set  ${\cal{Y}}_{\cal {A}}$.}
This assumption posits that the document has a correct answer span (or begin/end positions), but not every positive paragraph needs to have one. It further improves supervision quality ($\nearrow$), because for example, it allows the model to filter out the noise in paragraph two in \autoref{fig:noisy_sample}. Since the model is given a choice of any of the ${\cal{A}}$-consistent mentions, it has the capability to assign zero probability mass on the supervision-consistent mentions in that paragraph.

On the other hand, \texttt{H3} has lower coverage ($\searrow$) than \texttt{H1} and \texttt{H2}, because it provides a single positive example for the whole document, rather than one for each positive paragraph. It also reduces the strength of the supervision signal ($\searrow$), as the model now needs to figure out which mention to select from the larger document-level set  ${\cal{Y}}_{\cal {A}}$.

Note that we can only use \texttt{H3} coupled with a document-level model, because a paragraph-level model cannot directly tradeoff answers from different paragraphs against each other, to select a single answer span from the document. As with the other distant supervision hypotheses, span-based and position-based definitions of the possible consistent outcomes can be formulated. The  log-probabilities of these events are defined in row three of Table~\ref{tab:ds_assumptions}, when using $\sum$ for  $\Xi$.
\texttt{H3} was used by \newcite{kadlec-etal-2016-text} for cloze-style distantly supervised QA with recurrent neural network models.


The probability-space (paragraph vs.\ document-level) and the distant supervision assumption (\texttt{H1}, \texttt{H2}, and \texttt{H3}, each position or span-based) together define our interpretation of the distant supervision signal resulting in definitions of probability space outcomes consistent with the supervision.
Next, we define corresponding optimization objectives to train a model based on this supervision
and describe the inference methods to make predictions with a trained model.

%% file: sections/04_optimization_and_inference.tex
\section{Optimization and Inference Methods}
\label{sec:opt_and_infer}

For each distant supervision hypothesis, we 
maximize either 
the marginal log-likelihood of ${\cal A}$-consistent outcomes (MML)
or
the log-likelihood of the most likely outcome (HardEM).
The latter was found 
effective for weakly supervised tasks including QA and semantic parsing 
by \citet{Min-2019-EMNLP-hardem}.

\autoref{tab:ds_assumptions} shows the objective functions for all distant supervision assumptions,
each comprising a pairing of a distant supervision hypothesis (\texttt{H1}, \texttt{H2}, \texttt{H3}) and position-based vs.\ span-based interpretation.
The probabilities are defined according to the assumed probability space (paragraph or document). 
In the table, $\mathcal{K}$ denotes the set of all paragraphs in the document,
and $\mathcal{Y}^k$ denotes the set of weakly labeled answer spans for the paragraph $p_k$ (which can be $\{\emptyspan\}$ for paragraph-level models).
Note that span-based and position-based objective functions are  equivalent for \texttt{H1} because of the independence assumption, \ie
$ P_s(i^k, j^k) = P_b(i^k) P_e(j^k) $.


\noindent
{\bf Inference}:
Since the task is to predict an answer \emph{string} rather than a particular mention for a given question,  it is potentially beneficial to aggregate information across answer spans corresponding to the same string during inference. The score of a candidate answer string can be obtained as
$P_a (x) = \Xi_{(i, j) \in \mathcal{X}} P_s(i, j)$,
where $\mathcal{X}$ is the set of spans corresponding to the answer string $x$,
and $\Xi$ can be either $\sum$ or $\max$.\footnote{%
For inference with marginal ($\sum$) scoring, 
we use an approximate scheme where we only aggregate 
probabilities of candidates strings generated from a 20-best list of 
begin/end answer positions for each paragraph.} 
It is usually beneficial to match the training objective with the corresponding inference method,
\ie MML with marginal inference $\Xi=\sum$, and HardEM with max (Viterbi) inference $\Xi=\max$. \newcite{Min-2019-EMNLP-hardem} showed HardEM optimization was useful when using an \texttt{H2} span-level distant supervision assumption coupled with $\max$ inference,  but it is unclear whether this trend holds when $\sum$ inference is useful or other distant supervision assumptions perform better. We therefore study exhaustive combinations of probability space, distant supervision assumption, and training and inference methods.

%% file: sections/05_experiment.tex
\section{Experiments}
\label{sec:exp}

\input{sections/dataset_and_processing_implementation.tex}

\input{sections/span_vs_pos.tex}

\input{sections/distant_document_comparison.tex}

\input{sections/multi_obj_squad.tex}

\input{sections/test_eval.tex}

%% file: sections/dataset_and_processing_implementation.tex
\subsection{Data and Implementation}
Two datasets are used in this paper:  TriviaQA \cite{joshi-EtAl:2017:Long} in its Wikipedia formulation, and NarrativeQA (summaries setting) \cite{Kocisky2018TACL-narrativeqa}.
Using the same preprocessing as \newcite{clarkgardner} for TriviaQA-Wiki\footnote{\smaller\url{https://github.com/allenai/document-qa}},
we only keep the top 8 ranked paragraphs up to 400 tokens for each document-question pair for both training and evaluation.
Following \newcite{Min-2019-EMNLP-hardem}, for  NarrativeQA we define the possible answer string sets $\cal{A}$ using Rouge-L~\cite{Lin-2004-rouge} similarity with crouwdsourced abstractive answer strings. We use identical data preprocessing and the evaluation script provided by the authors.

In this work, we use the BERT-base model for text encoding and train our model with the default configuration as described in \cite{devlin-etal-2019-bert}, fine-tuning all parameters. We fine-tune for $3$ epochs on TriviaQA and $2$ epochs on NarrativeQA.

%% file: sections/span_vs_pos.tex
\subsection{Optimization and Inference for Latent Variable Models}
\label{ssec:span_vs_pos}

\begin{figure}[t]
    \centering
	\subfloat[TriviaQA F1]{
		\includegraphics[width=0.48\textwidth]
		{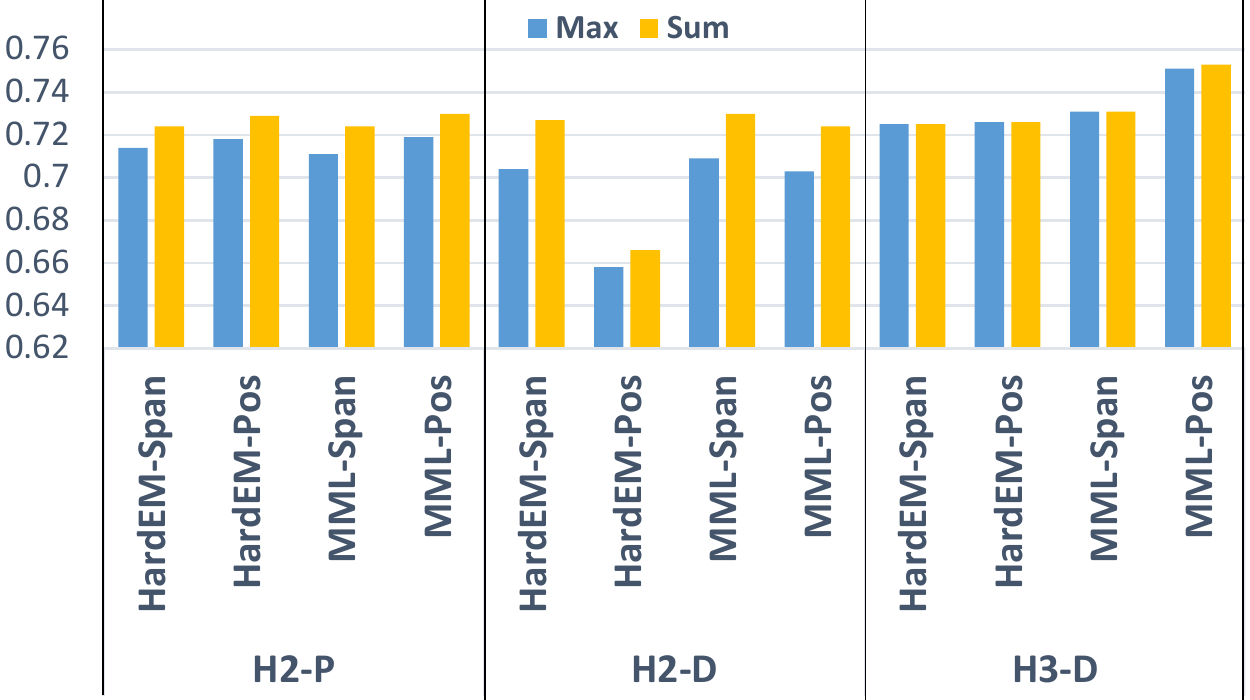}
		\label{fig:triviaqa_f1}
	} \\
	\subfloat[NarrativeQA Rouge-L]{
		\includegraphics[width=0.48\textwidth]
		{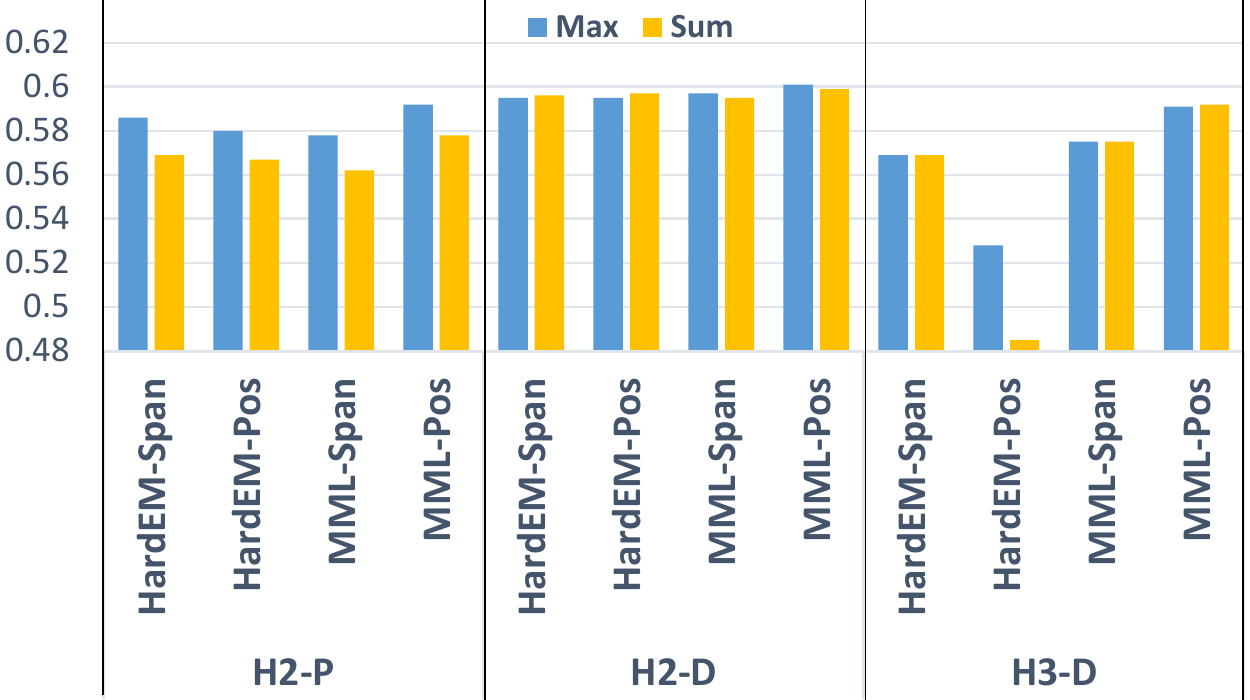}
		\label{fig:narrativeqa_rougel}
	}%
    \caption{Comparison of different optimization and inference choices grouped by distant supervision hypothesis based on dev set results for TriviaQA and NarrativeQA.}
	\label{fig:span_vs_pos}
\end{figure}

Here we look at the cross product of optimization  (HardEM vs MML) and inference (\textbf{Max} vs \textbf{Sum}) for all distant supervision assumptions that result in models with latent variables. We therefore exclude \texttt{H1} and look at the other two hypotheses, \texttt{H2} and \texttt{H3}, each coupled with a span-based (Span) or position-based (Pos) formulation and a paragraph-level (\texttt{P}) or a document level (\texttt{D}) probability space. The method used in \newcite{Min-2019-EMNLP-hardem} corresponds to span-based \texttt{H2}-\texttt{P} with HardEM training and \textbf{Max} inference.
The results are shown in \autoref{fig:span_vs_pos}.

First, we observe that inference with \textbf{Sum} leads to significantly better results on TriviaQA under \texttt{H2}-\texttt{P} and \texttt{H2}-\texttt{D}, and slight improvement under \texttt{H3}-\texttt{D}.
On NarrativeQA, inference with \textbf{Max} is better. We attribute this to the fact that correct answers often have multiple relevant mentions for TriviaQA (also see \S\ref{ssec:analysis}),
whereas for NarrativeQA this is rarely the case. Thus, inference with \textbf{Sum} in NarrativeQA could potentially boost the probability of irrelevant frequent strings.

Consistent with \cite{Min-2019-EMNLP-hardem}, we observe that 
span-based HardEM works better than span-based MML under \texttt{H2}-\texttt{P}, with a larger advantage on  NarrativeQA than on TriviaQA.
However, under \texttt{H2}-\texttt{D} and \texttt{H3}-\texttt{D}, span-based MML performs consistently better than span-based HardEM.
For position-based objectives, MML is consistently better than HardEM (potentially because HardEM may decide to place its probability mass on begin-end position combinations that do not contain mentions of strings in ${\cal A}$).
Finally, it can be observed that under each distant supervision hypothesis/probability space combination,
the position-based MML is always the best among the four objectives. Position-based objectives may perform better due to the independence assumptions for begin/end positions of the model we use and future work may arrive at different conclusions if position dependencies are integrated.
Based on this thorough exploration, we focus on experimenting with position-based objectives with MML for the rest of this paper.

%% file: sections/distant_document_comparison.tex
\subsection{Probability Space and Distant Supervision Assumptions}
\label{ssec:distant_doc_qa}


In this subsection, we compare probability space and distant supervision assumptions.
\autoref{tab:doc_qa_assumption_dev} shows the dev set results,
where the upper section compares paragraph-level models 
(\texttt{H1}-\texttt{P}, \texttt{H2}-\texttt{P}),
and the lower section compares document-level models
(\texttt{H1}-\texttt{D}, \texttt{H2}-\texttt{D}, \texttt{H3}-\texttt{D}).
The performance of models with both \textbf{Max} and \textbf{Sum} inference is shown.
We report F1 and Exact Match (EM) scores for TriviaQA,
and Rouge-L scores for NarrativeQA.

\begin{table}[t]
    \centering
    \small
    \begin{tabular}{c|c|c|c|c}
    \toprule
        \multirow{2}{*}{Objective} & 
        \multirow{2}{*}{Infer} & 
        \multicolumn{2}{c|}{ TriviaQA} & {NarrativeQA}\\
        \cmidrule{3-5}
        & & F1 & EM & Rouge-L\\
        \midrule
        \multicolumn{5}{c}{Paragraph-level Models} \\
        \midrule
        \multirow{2}{*}{\texttt{H1-P}} & Max & 67.9 & 63.3 & 55.3\\
        & Sum & 70.4 & 66.0 & 53.6 \\
        \midrule
        \multirow{2}{*}{\texttt{H2-P}} & Max & 71.9 & 67.7 & 59.2 \\
        & Sum & 73.0 & 69.0 & 57.8 \\
        \midrule
        \multicolumn{5}{c}{Document-level Models} \\
        \midrule
        \multirow{2}{*}{\texttt{H1-D}} & Max & 55.8 & 51.0 &  59.4\\
        & Sum & 65.2 & 61.2 & 59.1\\
        \midrule
        \multirow{2}{*}{\texttt{H2-D}} & Max & 70.3 & 66.2 & 60.1\\
        & Sum & 72.4 & 68.4 & 59.9\\
        \midrule
        \multirow{2}{*}{\texttt{H3-D}} & Max & 75.1 & 70.6 & 59.1\\
        & Sum & 75.3 & 70.8 & 59.2\\
    \bottomrule
    \end{tabular}
    \caption{Comparison of distant supervision hypotheses using 
    MML-Pos
    objectives on TriviaQA and NarrativeQA dev sets.
    }
    \label{tab:doc_qa_assumption_dev}
\end{table}

For TriviaQA, \texttt{H3}-\texttt{D} achieves significantly better results than other formulations. Only \texttt{H3}-\texttt{D} is capable of ``cleaning'' noise from positive paragraphs that don't have a correct answer (e.g. paragraph two in \autoref{fig:noisy_sample}), by deciding which ${\cal{A}}$-consistent mention to trust.
The paragraph-level models \texttt{H1}-\texttt{P} and \texttt{H2}-\texttt{P} 
outperform their corresponding document-level counterparts \texttt{H1}-\texttt{D} and \texttt{H2}-\texttt{D}. This may be due to the fact that without \texttt{H3}, and without predicting \emptyspan, \texttt{D} models do not learn to detect irrelevant paragraphs.

Unlike for TriviaQA, \texttt{H2}-\texttt{D} models achieve the best performance for NarrativeQA.
We hypothesize this is due to the fact that positive paragraphs that don't have a correct answer are very rare in NarrativeQA (as summaries are relatively short and answer strings are human-annotated for the specific documents). Therefore, \texttt{H3} is not needed to clean noisy supervision, and it is not useful since it also leads to a reduction in the number of positive examples (coverage) for the model. Here,  document-level models always improve over their paragraph counterparts, by learning to calibrate paragraphs directly against each other. 


%% file: sections/multi_obj_squad.tex
\subsection{Multi-Objective Formulations and Clean Supervision}
\label{ssec:multi_obj_squad}

Here we study two methods to further improve weakly supervised QA models.
First, we combine two distant supervision objectives in a multi-task manner, \ie
\texttt{H2}-\texttt{P} and \texttt{H3}-\texttt{D} for TriviaQA,
and \texttt{H2}-\texttt{P} and \texttt{H2}-\texttt{D} for NarrativeQA,
chosen based on the results in \S\ref{ssec:distant_doc_qa}. \texttt{H2} objectives have higher coverage than \texttt{H3} while being more susceptible to noise. Paragraph-level models have the advantage of learning to score irrelevant paragraphs (via \emptyspan\ outcomes). Note that we use the same parameters for the two objectives and the multi-objective formulation does not have more parameters and is no less efficient than the individual models.
Second, we use external clean supervision from SQUAD 2.0 \cite{rajpurkar-etal-2018-know}
to train the BERT-based QA model for 2 epochs.
This model matches the \texttt{P} probability space and is able to detect both \emptyspan\ and extractive answer spans.
The resulting network is used to initialize the models for TriviaQA and NarrativeQA.
The results are shown in \autoref{tab:doc_qa_multi_obj_clean_dev}.

\begin{table}[]
    \centering
    {\small
    \begin{tabular}{c|c|c|c|c|c}
    \toprule
        \multirow{2}{*}{Objective} 
        & \multirow{2}{*}{Clean}       
        & \multirow{2}{*}{Infer}       
        & \multicolumn{2}{c|}{ TriviaQA} & {NarrativeQA}\\
        \cmidrule{4-6}
        & & & F1 & EM & Rouge-L \\
        \midrule
        \multicolumn{6}{c}{Single-objective} \\
        \midrule
        \multirow{4}{*}{\textbf{Par}} & \multirow{2}{*}{X} & Max & 71.9 & 67.7 & 59.2\\
        & 
        & Sum & 73.0 & 69.0 & 57.8 \\
        \cmidrule{2-6}
                                         & \multirow{2}{*}{\checkmark} & Max & 74.2 & 70.1 & 61.7\\
        & 
        & Sum & \textbf{74.9} & \textbf{70.9} & \textbf{61.7} \\
        \midrule
        \multirow{4}{*}{\textbf{Doc}} & \multirow{2}{*}{X} & Max & 75.1 & 70.6 & 60.1\\
        & 
        & Sum & 75.3 & 70.8 & 59.9 \\
        \cmidrule{2-6}
            & \multirow{2}{*}{\checkmark} & Max & 75.5 & 70.8 & 62.8\\
        & 
        & Sum & \textbf{75.5} & \textbf{70.9} & \textbf{62.9} \\
        \midrule
        \multicolumn{6}{c}{Multi-objective} \\
        \midrule
        \multirow{4}{*}{\shortstack{\textbf{Par}\\+\\\textbf{Doc}}} 
        & \multirow{2}{*}{X} & Max & 75.6 & 71.2 & 60.5\\
        & 
        & Sum & 75.9 & 71.6 & 60.5 \\
        \cmidrule{2-6}
         & \multirow{2}{*}{\checkmark} & Max & 75.8 & 71.2 & 63.0\\
        & 
        & Sum & \textbf{76.2} & \textbf{71.7} & \textbf{63.1} \\
    \bottomrule
    \end{tabular}}
    \caption{Dev set results comparing multi-objectives and clean supervison.
    \text{\checkmark} indicates the QA model is pre-trained on SQUAD.}
    \label{tab:doc_qa_multi_obj_clean_dev}
\end{table}


It is not surprising that 
using external clean supervision improves model performance (e.g.\ \cite{min-seo-hajishirzi:2017:Short}).
We note that, interestingly, this external supervision narrows the performance gap between paragraph-level and document-level models, and reduces the difference between the two inference methods.

Compared with their single-objective components, multi-objective formulations improve performance on both TriviaQA and NarrativeQA.

%% file: sections/test_eval.tex
\subsection{Test Set Evaluation}
\label{ssec:test_val}

\begin{table}[t]
    \centering
    \resizebox{0.47\textwidth}{!}{
    \begin{tabular}{l|c | c | c |c}
    \toprule
    \multicolumn{5}{c}{TriviaQA Wiki}\\
    \midrule
                 & \multicolumn{2}{c|}{Full} & \multicolumn{2}{c}{Verified} \\
                 \cmidrule{2-5}
                 &  F1 &  EM & F1 & EM \\
    \midrule
        Ours (\textbf{H2-P+H3-D})  & \textbf{76.3} & \textbf{72.1} & \textbf{85.5} & \textbf{82.2} \\
        \quad \textit{w/o} SQUAD  & 75.7 & 71.6 & 83.6 & 79.6 \\
        \hline
        \cite{wang-etal-2018-multi-granularity} & 71.4 &  66.6  & 78.7 & 74.8\\
        \cite{clarkgardner}                     & 68.9 &  64.0  & 72.9 & 68.0 \\
        \cite{Min-2019-EMNLP-hardem}            & 67.1 &   --     & --   &  -- \\
        \midrule
        \multicolumn{5}{c}{NarrativeQA Summary}\\
    \midrule
        & \multicolumn{4}{c}{Rouge-L}\\
    \midrule
        Ours (\textbf{H2-P+H2-D}) & \multicolumn{4}{c}{\textbf{62.9}}\\
           \quad\textit{w/o} SQUAD  & \multicolumn{4}{c}{60.5}\\
        \midrule
        \cite{nishida-etal-2019-multi} & \multicolumn{4}{c}{59.9}\\
             \quad\textit{w/o} external data 
             & \multicolumn{4}{c}{54.7}\\
        \midrule
        \cite{Min-2019-EMNLP-hardem} & \multicolumn{4}{c}{58.8}\\
    \bottomrule
    \end{tabular}
    }
    \caption{Test set results on TriviaQA Wiki and NarrativeQA Summaries.
    ``\textit{w/o} SQUAD'' refers to our best model without pretraining on SQUAD 2.0.
    ``\textit{w/o} external data'' refers to the model 
    from \cite{nishida-etal-2019-multi} without using MS MARCO data \cite{msmarco2018v3}.}
    \label{tab:triviaqa_leaderboard}
\end{table}

\autoref{tab:triviaqa_leaderboard} reports test set results on TriviaQA 
and NarrativeQA for our best models, in comparison to recent state-of-art (SOTA) models.
For TriviaQA, we report F1 and EM scores on the full test set and the verified subset.
For NarrativeQA, Rouge-L scores are reported.

Compared to recent TriviaQA SOTA \cite{wang-etal-2018-multi-granularity}, 
our best models achieve $4.9$ F1 and $5.5$ EM improvement on the full test set,
and 6.8 F1 and 7.4 EM improvement on the verified subset.
On the NarrativeQA test set, we improve  Rouge-L by 3.0
over \cite{nishida-etal-2019-multi}.
The large improvement, even without additional fully labeled data, demonstrates the importance of selecting an appropriate probability space and interpreting the distant-supervision in a way cognizant of the properties of the data, as well as selecting a strong optimization and inference method. 
With external fully labeled data to initialize the model,
performance is further significantly improved.



%% file: sections/06_analysis.tex
\subsection{Analysis}
\label{ssec:analysis}
In this subsection, we carry out analyses to study the relative performance of paragraph-level and document-level models, depending on the size of answer string set $|\mathcal{A}|$ and the number of $\cal{A}$-consistent spans, which are hypothesized to correlate with label noise. We use the TriviaQA dev set
and the best performing models, \ie \texttt{H2}-\texttt{P} and \texttt{H3}-\texttt{D} with \textbf{Sum} inference.

We categorize examples based on the size of their answer string set, $|\mathcal{A}|$, and 
the size of their corresponding set of ${\cal{A}}$-consistent spans, $|\mathcal{I}|$.
Specifically, we divide the data into 4 subsets and report performance separately on each subset, 
as shown in \autoref{tab:partition_analysis_triviaqa}.
In general, we expect $\mathcal{Q}^{sl}$ and $\mathcal{Q}^{ll}$ to be noisier
due to the larger $\mathcal{I}$,
where $\mathcal{Q}^{sl}$ potentially includes many irrelevant mentions 
while $\mathcal{Q}^{ll}$ likely contains more incorrect answer strings (false aliases).
We can observe that the improvement is more significant for these noisier subsets,
suggesting document-level modeling is crucial for handling both types
of label noise.

\begin{table}[t]
    \centering
    \small
    \begin{tabular}{ccccccc}
    \toprule
        Subset & $|\mathcal{A}|$ & $|\mathcal{I}|$ & Size
        & H2-P & H3-D
        & $\Delta$ \\
        \midrule
        $\mathcal{Q}^{ss}$ & $= 1$ & $\le 5$ & 2585 
        & 66.8 & 67.4 & 0.6\\
        $\mathcal{Q}^{ls}$ & $> 1$ & $\le 5$ & 853 
        & 68.7 & 70.1 & 1.4\\
        $\mathcal{Q}^{sl}$ & $= 1$ & $> 5$ & 1149 
        & 82.0 & 84.9 & 2.9\\
        $\mathcal{Q}^{ll}$ & $> 1$ & $> 5$ & 3034 
        & 86.3 & 88.4 & 2.1\\
    \bottomrule
    \end{tabular}
    \caption{
    F1 scores on 4 subsets of TriviaQA dev, grouped by the size of their answer string sets ${\cal{A}}$ and corresponding set of possible mentions ${\cal{I}}$.
    $\Delta$ indicates the improvement from \texttt{H2}-\texttt{P} to \texttt{H3}-\texttt{D}.}
    \label{tab:partition_analysis_triviaqa}
\end{table}

%% file: sections/07_related.tex
\section{Related Work}

Distant supervision has been successfully used for decades for information extraction tasks such as entity tagging and relation extraction \cite{Craven:1999:CBK:645634.663209,mintz2009distant}.
Several ways have been proposed to learn with DS, e.g., 
 multi-label multi-instance learning~\cite{Surdeanu:2012:MML:2390948.2391003},
assuming at least one supporting evidence~\cite{hoffman2011knowledge},
integration of label-specific priors~\cite{ritter2013modeling},
and adaption to shifted label distributions~\cite{ye-etal-2019-looking}.

Recent work has started to explore distant supervision to scale up QA systems,
particularly for open-domain QA
where the evidence has to be retrieved rather than given as input.
Reading comprehension (RC) with evidence retrieved from information retrieval systems 
establishes a weakly-supervised QA setting due to the noise in the heuristics-based span labels
\cite{drqa,joshi-EtAl:2017:Long,Dunn2017search_qa,Dhingra2017Quasar}. 
One line of work jointly learns RC and evidence ranking using either a pipeline system \cite{Wang2018aaai,lee-etal-2018-ranking,kratzwald-feuerriegel-2018-adaptive}
or an end-to-end model \cite{lee-etal-2019-latent}.

Another line of work focuses on improving distantly-supervised RC models by developing learning methods and model architectures that can better use noisy labels.
\citet{clarkgardner} propose a paragraph-pair ranking objective, which has components of both our \texttt{H2-P} and \texttt{H3-D} position-based formulations. They don't explore multiple inference methods or combinations of objectives and use less powerful representations.
In \cite{lin-etal-2018-denoising}, a coarse-to-fine model is proposed to handle label noise by aggregating information from relevant paragraphs and then extracting answers from selected ones.
\citet{Min-2019-EMNLP-hardem} propose a hard EM learning scheme which we included in our experimental evaluation.

Our work focuses on examining probabilistic assumptions for document-level extractive QA.
We provide a unified view of multiple methods in terms of their probability space and distant supervision assumptions and evaluate the impact of their components in combination with optimization and inference methods. To the best of our knowledge, the three DS hypotheses along with position and span-based interpretations have not been formalized and experimentally compared on multiple datasets. In addition, the multi-objective formulation is new.

%% file: sections/08_conclusion.tex
\section{Conclusions}
\label{sec:conclusion}
In this paper, we demonstrated that the choice of probability space and interpretation of the distant supervision signal for document-level QA have a large impact, and that they interact. Depending on the properties of the data, different configurations are best, and a combined multi-objective formulation can reap the benefits of its constituents.

A future direction is to extend this work to question answering tasks that require reasoning over multiple documents, e.g., open-domain QA. 
In addition, the findings may generalize to other tasks,
e.g., corpus-level distantly-supervised relation extraction.